%% file: main.tex
\documentclass[sigconf,nonacm]{acmart}
\AtBeginDocument{%
  }

\usepackage{xspace}

\newcommand{\etal}{{\em et al. }}
\newcommand{\BfPara}[1]{{\noindent {\bf #1}}}

\usepackage{pifont}
\newcommand{\cmark}{\ding{51}}%
\newcommand{\xmark}{\ding{55}}%

\usepackage{multirow}
\usepackage{paralist} 

\usepackage{threeparttable}
\usepackage{booktabs}

\newcommand{\Workname}{{\textit{GuardedTuning}}\xspace}

\usepackage{titlesec}

\titlespacing*{\subsection}{0pt}{*0.5}{*0.5}
\titlespacing*{\section}{0pt}{*1.0}{*0.5}

\begin{document}

\title{Navigating the Designs of Privacy-Preserving Fine-tuning for Large Language Models}

\author{Haonan Shi}
\email{haonan.shi3@case.edu}
\affiliation{%
  \institution{Case Western Reserve University}
  \country{}
}

\author{Tu Ouyang}
\email{tu.ouyang@case.edu}
\affiliation{%
  \institution{Case Western Reserve University}
  \country{}
}

\author{An Wang}
\email{an.wang@case.edu}
\affiliation{%
  \institution{Case Western Reserve University}
  \country{}
}


\begin{abstract}
 Instruction tuning has proven effective in enhancing Large Language Models' (LLMs) performance on downstream tasks. 
 However, real-world fine-tuning faces inherent conflicts between model providers' intellectual property protection, clients' data privacy requirements, and tuning costs.
 While recent approaches like split learning and offsite tuning demonstrate promising architectures for privacy-preserving fine-tuning, there is a gap in systematically addressing the multidimensional trade-offs required for diverse real-world deployments.
 We propose several indicative evaluation metrics to guide design trade-offs for privacy-preserving fine-tuning and a series of example designs, collectively named \Workname; they result from novel combinations of system architectures with adapted privacy-enhancement methods and emerging computation techniques.
 Each design represents distinct trade-offs across model utility, privacy guarantees, and costs.
 Experimental results demonstrate that these designs protect against data reconstruction attacks while maintaining competitive fine-tuning performance.

\end{abstract}





\maketitle

\input{introduction}

\input{relatedworks}
\input{methodology}

\input{experiment}

\input{conclusion}


\bibliographystyle{abbrv}
\bibliography{reference}

\end{document}

%% file: introduction.tex
\section{INTRODUCTION}
Large Language Models (LLMs)~\cite{zhang2022opt,touvron2023llama} have demonstrated exceptional performance across a wide range of natural language processing tasks. 
Through pretraining on massive text corpora, these models learn rich language representations and knowledge that exhibit strong generalization capabilities across various downstream tasks. 
Although LLMs typically demonstrate reasonable zero-shot learning abilities, research shows that fine-tuning on specific tasks can significantly improve model performance on these tasks~\cite{wei2021finetuned}.
Fine-tuning the LLMs with domain-specific data has emerged as an important paradigm to adapt pre-trained LLMs to better suit some domain-specific tasks.

Due to substantial computational investments, model providers consider these LLMs proprietary assets and want to guard against leakage. 
Conversely, even though some model providers offer fine-tuning services through APIs (e.g., OpenAI fine-tuning APIs~\footnote{https://platform.openai.com/docs/guides/text-generation}), it requires the client to upload their private data to the model provider, fine-tuning clients might want to safeguard the fine-tuning data even against model providers due to the sensitivity of the data or regulation requirements. 
Overall, the requirements of real-world fine-tuning solutions are diverse and subject to sometimes conflicting participating parties' concerns, particularly regarding privacy and budget.

To address these privacy conflicts between fine-tuning participants,
Recent privacy-preserving approaches like split learning~\cite{vepakomma2018split, wang2023privatelora} and offsite tuning~\cite{xiao2023offsite} propose new system architectures that partition the training process between providers and users, where only model intermediate values or compressed model components are exchanged rather than raw input data. Different system architectures enable distinct trade-offs that fit some fine-tuning requirements but not others. In addition, privacy-enhancement methods, some of which choose to add noise with respect to different optimization objectives to either input or intermediate data, further broaden the privacy-preserving fine-tuning solution design space.

We advocate evaluating these emerging privacy-preserving fine-tuning designs using multiple metrics but not any single one. We propose using metrics in three categories, utility, privacy (of each participating party), and tuning cost, for both design evaluation and to help navigate the design trade-offs necessary to meet diverse real-world requirements. We acknowledge that different system architectures, combined with different privacy-enhancement methods, could result in various solutions with distinct utility, privacy, and cost characteristics.

To showcase how different design trade-offs impact the metrics and hence help navigate the design space, we produce a series of example designs for privacy-preserving fine-tuning, collectively named \Workname, from novel combinations of system architectures, adapted privacy-enhancement methods and emerging computation techniques. We articulate the trade-offs underneath these designs and their differences from existing ones in the literature in section~\ref{sec:our_designs} and present some preliminary quantitative results on the evaluation metrics these designs achieve in section~\ref{sec:evals}.
Our experiment results demonstrate that \textit{GuardedTuning} effectively protects both server's model weights and client's data privacy, reducing data reconstruction attacks effectiveness to below 50\% and decreasing communication cost by up to 73.7\%, while the degrade of fine-tuned model utility is <1.5\% in almost all of the cases.

%% file: methodology.tex
\section{THREAT MODEL AND DESIGN TRADE-OFFS}
In our privacy-preserving fine-tuning setting, we consider an honest-but-curious model provider (also referred to as \textit{server}) that has access to the pre-trained model weights. Fine-tuning client provides data for fine-tuning training and inferences. The fine-tuning training and inference could be split between these two parties.
While faithfully following the fine-tuning protocol, the server can not be fully trusted and might try to look into the client's data by performing data reconstruction attacks(DRAs) during either training or inference phases using the activations or gradients transmitted from client. 
Conversely, the model provider also needs to protect their intellectual property.
So they cannot hand over the pre-trained model to the client.

There is a vast design space to produce a specific fine-tuning design that satisfies the privacy concerns of both the server and the client, as well as other requirements arising from, for instance, the utility of the tuned model and operation budgets.
We propose to use the following metrics to evaluate different solutions and leverage the metrics as guidance to arrive at the right balance when making concrete design choices for a specific set of requirements: 
\begin{inparaenum} 
    \item The utility of the fine-tuned model;   
    \item The privacy of clients' data for fine-tuning training;   
    \item The privacy of clients' data for inference with fine-tuned model;
    \item The server's model privacy;
    \item Communication cost associated with fine-tuning.
\end{inparaenum}

Our concrete designs below show how to navigate the trade-offs among these metrics.

\section{GUARDEDTUNING ARCHITECTURES}
\label{sec:our_designs}

\subsection{\textbf{Online GuardedTuning}}
\begin{figure}[!htbp]
    \centering
    \captionsetup{justification=justified}
    \includegraphics[width=1\linewidth]{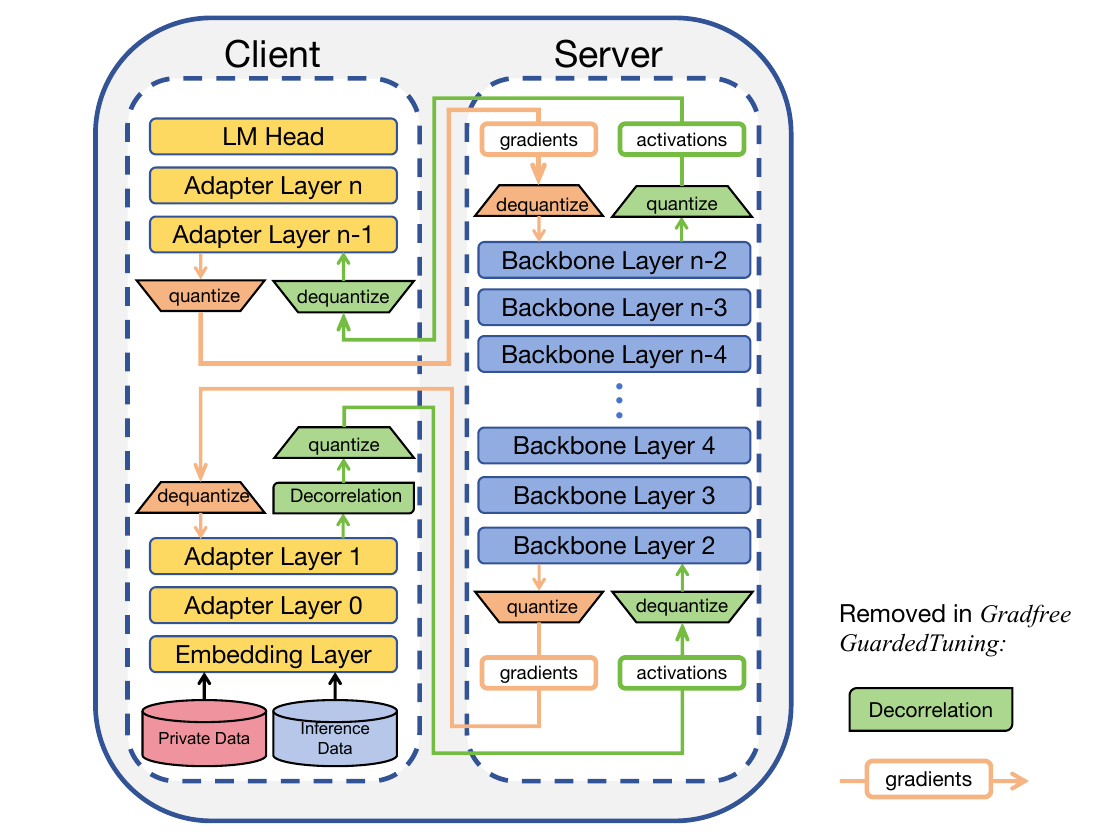}
    \caption{Architectures for \textit{Online GuardedTuning} and \textit{Gradfree GuardedTuning}.}
    \label{fig:online}
\end{figure}

Inspired by the split learning, we propose \textit{Online GuardedTuning}, as shown in Figure \ref{fig:online}. 
In this design, the server distributes several bottom and top transformer layers of a pre-trained LLM to the client for collaborative fine-tuning. 
We call the layers clients hold adapter layers, and those servers kept as backbone layers. 
During fine-tuning training, the client and server exchange activations for forward pass and gradients for backward propagation.
Client data privacy is enhanced by avoiding plain text transfers to the server, though this introduces additional communication cost during both fine tuning and inference.

This design has a utility-cost trade-off: while adapter layers in the client always require fine-tuning, the server can choose to fine-tune the backbone layers or not—the second choice results in lower overall cost but with a reduced model utility. Prior research~\cite{wu2024fedbiot,xiao2023offsite} suggests that fine-tuning a few selected layers of the model can still yield reasonable performance, slightly lower than fully fine-tuning.

Through the activations uploaded by clients, the server can reconstruct clients' data via DRAs~\cite{song2020information, chen2024unveiling} with >90\% attack effectiveness, rendering the fine-tuning design useless as client data is close to having no protection.
Existing solutions like PrivateLoRA\cite{wang2023privatelora}, despite retaining embedding layer, head layer and low-rank adapters locally, remain vulnerable to server-side DRAs due to unprotected activation transmission.

To defend against such attacks and realize better client data privacy, we introduce two novel privacy-enhancing methods that can be composed into several \textit{GuardedTuning} designs. These methods help partially dwarf such DRAs at the cost of a small amount of extra computation for data perturbation.

First, We apply a ~\textit{distance decorrelation} technique, adapted from the Nopeek~\cite{vepakomma2020nopeek}, to transform activations output from the input adapter layers. Vepakomma \etal identify that the distance correlation $dCor(x, \theta(x))$ between input data $x$ and activations $\theta(x)$ indicates the risk of input data being inverted from intermediary representations, they incorporate $dCor(x, \theta(x))$ as a regularization term in the training loss function. 
By minimizing $dCor(x, \theta(x))$ during training, they could mitigate DRAs. 
Unlike Nopeek, which uses Euclidean distance in computing distance correlation between input data $x$ and activations $\theta(x)$, we instead calculate the distance correlation between input embeddings $emb(x)$ and activations $\theta(x)$ using cosine distance, which achieves better performance-privacy trade-off for LLMs.
Cosine distance provides a more appropriate representation of distances in the high-dimensional embedding space in LLMs.
After incorporating the decorrelation operation, the fine-tuning loss function becomes:
\begin{equation}
    L = L(f(x), y) + \lambda dCor(emb(x), \theta(x))
\label{eq:decorrelation}
\end{equation}
where $f$ represent the LLM, $\theta$ denotes the input adapter layers, $dCor()$ denotes the distance correlation function\cite{vepakomma2020nopeek}, and $\lambda$ represents the regularization weight. 

However, adding a regularization term alone cannot protect fine-tuning data in early training stages. To address this limitation, we additionally apply a quantization-based technique that protects the beginning stage of the training. During transmission, the sender quantizes the inlier values of activations and gradients while preserving outliers:
\begin{equation}
Q(A) = \begin{cases}
\text{round}\left(\frac{A - \min(A)}{P_{p}(A) - \min(A)} \cdot (2^b - 1)\right) & \text{if } A \leq P_{p}(A) \\
A & \text{if } A > P_{p}(A)
\end{cases}
\label{eq:quantization}
\end{equation}
\begin{equation}
\hat{A} = \begin{cases}
Q(A) \cdot s + \min(A) & \text{if bit-width}(Q(A)) = b \\
Q(A) & \text{if bit-width}(Q(A)) = 32
\end{cases}
\end{equation}
where $s = \frac{P_{p}(A) - \min(A)}{2^b - 1}$ is the scaling factor transmitted with $\min(A)$ and quantized values $Q(A)$. The quantization bits $b$ and the $p$-th percentile threshold are determined locally by the sender.  Our experiments show that quantization introduces controlled noise that helps realize better privacy protection without leading to significant model performance degradation after fine-tuning. 


We note that quantization also helps reduce communication costs since the transmitted data is smaller in size than split learning designs.
Additional techniques could be included in this design to reduce communication costs further. 
For instance, by leveraging emerging trusted computing technologies, like hardware-assisted secure GPU sharing~\cite{nvidia_gpu_secure}, or confidential computing techniques that enable two parties sharing the same host machine with privacy attestation by cloud vendors~\cite {azure_confidential}, it is possible to run separated components of the fine-tuning design on the same machine, even on the same GPU, with similar privacy protection as running them on different machines. While the communication volume is the same, the time of achieving a similar amount of communication significantly improves if employing these techniques, such as NVLink.

\subsection{\textbf{Gradfree GuardedTuning}}
Although \textit{Online GuardedTuning} offers substantial decreases in the DRAs attack effectiveness as shown in our evaluations in section \ref{sec:evals}, we note that privacy attacks like gradient matching-based DRAs~\cite{chen2024unveiling, deng2021tag} can still marginally improve DRA's performance.
We introduce a design called \textit{Gradfree GuardedTuning}, which favors better client data privacy than \textit{Online GuardedTuning} at the cost of a slight decrease in model utility.

As shown in Figure \ref{fig:online}, in \textit{Gradfree GuardedTuning}, we choose to fine-tune only the output adapter; by doing so, it removes the gradient backpropagation between the client and server required in \textit{online GuardedTuning}.
Removing the gradient backpropagation provides full defense against gradient matching-based DRAs. 
Although the absence of gradient transmission disallows some other privacy-enhancing methods like decorrelation operations, we still can apply method like quantization on activations in this design, this method provides substantive defense to several DRAs.

\subsection{\textbf{Offline GuardedTuning}}
\begin{figure}[!htbp]
    \centering
    \captionsetup{justification=justified}
    \includegraphics[width=0.89\linewidth]{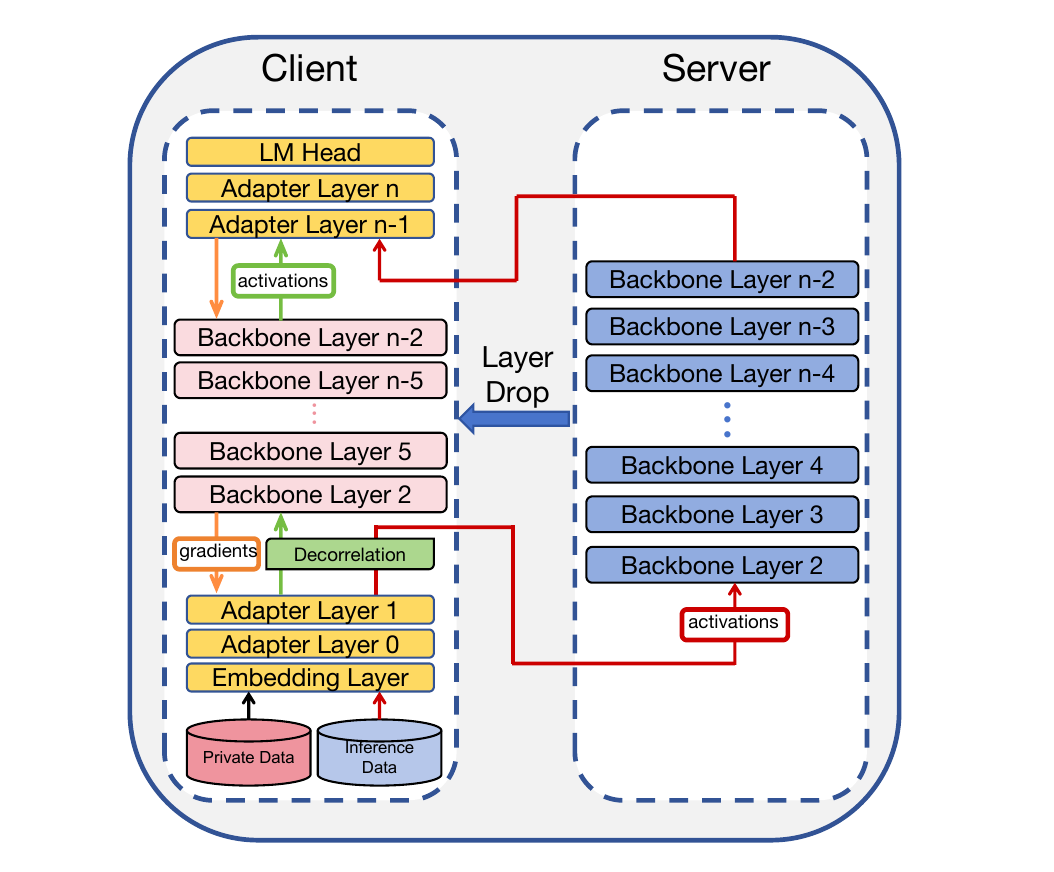}
    \caption{Architecture for \textit{Offline GuardedTuning}.}
    \label{fig:offline}
\end{figure}

To realize better client data privacy, we design \textit{offline GuardedTuning}, an enhancement to offsite-tuning~\cite{xiao2023offsite}. This design is as shown in Figure \ref{fig:offline}, 
Similar to offsite tuning, in our design, during fine-tuning training, the server provides clients with adapter layers and an emulator constructed through uniform layers dropping from the backbone. 
This emulator enables clients to simulate the backbone layers and run fine-tuning training entirely within the client, thereby protecting fine-tuning data from server-side data reconstruction attacks during training.
The design trade-off is to sacrifice the sever model privacy to a certain degree (because the emulator leaks information about the base model) in exchange for complete client training data privacy.

In offsite tuning, the client is supposed to transfer their fine-tuned input and output layers to the server and run the entire inference there, leaving the inference-time client data vulnerable. 
To address this privacy limitation, in \textit{offline guardedtuning}, we choose to let the client keep the input and output adapter layers, only transferring activations to the server for inference.
Furthermore, we co-train decorrelation operators during the fine-tuning training phase and use the decorrlation-tuned input adapter layers during inference for activation values, to protect inference data against DRAs. By adding decorrelation operation, our design enhances client inference data privacy with a small amount of data perturbation cost.

Xiao \etal ~\cite{xiao2023offsite} observe that the emulator, coupled with fine-tuned adapters, their inference performance may be significantly weaker than the base model's zero-shot.
Therefore, in our design, we choose to use the backbone layers in the server instead of the emulator for inference, a similar design choice in offsite learning.
This low emulator inference performance might indicate a lower knowledge capacity in the emulator and, thus, a lower risk of allowing the client to reverse-engineer the base model using an emulator or leverage it for other downstream tasks.

%% file: experiment.tex
\section{EVALUATIONS}
\label{sec:evals}
In this section, we evaluate the three proposed \textit{GuardedTuning} designs and show how their trade-offs affect the utility, privacy, and cost metrics.

\subsection{Experiment Setup}
\BfPara{Datasets.} We evaluate \textit{GuardedTuning} on four reasoning task datasets: ARC-Easy~\cite{clark2018think} and OpenBookQA~\cite{mihaylov2018can} for multiple-choice scientific reasoning, PIQA~\cite{bisk2020piqa} for binary physical commonsense reasoning, and WebQuestions~\cite{berant2013semantic} for open-domain question answering.

\BfPara{Evaluated Models.} 
We conducted experiments using OPT-1.3B and OPT-6.7B models~\cite{zhang2022opt}. 

\BfPara{Experiment Configurations.}
In experiments using OPT-1.3B, the client holds the first and last two layers as adapters with the server keeping 20 backbone layers (8 layers for the backbone emulator), while in experiments involving OPT-6.7B, 3 adapter layers for both input and output, and 26 backbone layers (14 layers for the emulator). 
During fine-tuning, we set the decorrelation weight $\lambda$ to 5 in Equation \ref{eq:decorrelation}, employ 8-bit quantization for transmission, and use the 99th percentile as the quantization threshold in Equation \ref{eq:quantization}.

\BfPara{Metrics.}
We measure \textbf{\textit{client data privacy}} through state-of-the-art DRAs (BiSR(b) and BiSR(f))~\cite{chen2024unveiling} during both fine-tuning and inference phases. Attack effectiveness is quantified using ROUGE-L F1 scores, with higher scores indicating greater privacy leakage. DRAs are conducted via three stages: inverter training with distributed adapters, initial reconstruction using server activations, and optimization through activations/gradients. 
We conduct evaluations every 100 training steps, averaging results across 5 independent batches.
For the designs that involve gradient exchange (split learning-based fine-tuning, \textit{online GuardedTuning}), we apply gradient-based BiSR(b) during fine-tuning. Otherwise, we employ BiSR(f) to attack the activations from forward propagation during fine-tuning and inference.
For \textit{\textbf{server model privacy}}, we use the number of shared model layers as a quantitative proxy metric.
\textit{\textbf{Model utility}} is measured as task-specific accuracy using lm-evaluation-harness~\cite{lm-evaluation-harness}.
To evaluate \textit{\textbf{communication cost}}, we quantify the data transmission volume between client and server during the fine-tuning process.

\BfPara{Baseline.}
We use traditional split learning-based fine-tuning(SL tuning)~\cite{vepakomma2018split} and offsite tuning\cite{xiao2023offsite} as our baseline.

\subsection{Experiment Results}
We evaluate all the \textit{GuardedTuning} designs on OPT-1.3B and OPT-6.7B models. 
Table \ref{tab:opt-1.3b} shows OPT-1.3B results on the utility and privacy trade-off, that all designs reach reasonable utility-privacy balance: all surpass OPT-1.3B's zero-shot task accuracy, achieve slightly lower accuracy than SL tuning baseline that has no defense to state-or-the-art DRAs while significantly reducing the server's DRA attack effectiveness to less than 50\% (a decrease of at least 40\% from the baseline) during both fine-tuning and inference phases. 
Additionally, our quantization-dequantization mechanism reduces communication cost by up to 73.7\% during fine-tuning.
\begin{table}[!htbp]
\centering
\scriptsize
\begin{threeparttable}
\caption{Utility and privacy trade-offs on OPT-1.3B.}
\label{tab:opt-1.3b}
\begin{tabular}{lcccccccc}
\toprule
\multirow{2}{*}{Designs} & 
\multirow{2}{*}{Metrics} & 
\multicolumn{4}{c}{OPT-1.3B} \\ 
\cmidrule(lr){3-6}
 & & 
 OpenbookQA & ARC-Easy & PIQA & WebQs \\ 
\midrule
Zero Shot
 & Accuracy
 & 23.4\% & 56.9\% & 71.6\% & 4.6\%  \\
\midrule
SL tuning
 & Accuracy
 & 31.4\% & 61.3\% & 75.2\% & 31.2\%  \\
({\color{red}{2}}-20-{\color{red}{2}}) & Fine-tune privacy
 & 89.4 & 94.4 & 86.3 & 88.2 \\
 & Inference privacy
 & 84.6 & 92.4 & 83.1 & 87.5  \\
 & Communication cost(GB)
 & 15.2 & 13.3 & 326.1 & 14.1\\
\midrule
offsite tuning
 & Accuracy
 & 29.1\% & 59.4\% & 74.5\% & 27.2\%  \\
  ({\color{red}{2}}-20({\color{red}{8}})-{\color{red}{2}}) & Fine-tune privacy
 & N/A & N/A & N/A & N/A \\
 & Inference privacy
 & 100 & 100 & 100 & 100 \\
 & Communication cost(GB)
 & 3.5 & 3.5 & 3.5 & 3.5 \\
\midrule
online GT
 & Accuracy
 & 30.7\% & 60.2\% & 73.9\% & 32.2\%  \\
 ({\color{red}{2}}-20-{\color{red}{2}}) & Fine-tune privacy
 & 37.7  & 41.1  & 28.0 & 38.7 \\
 & Inference privacy
 & 30.1  & 30.5  & 24.8 & 21.2  \\
  & Communication cost(GB)
 & 4.6 & 4.1 & 85.7 & 4.3\\
\midrule
 Gradfree GT
 & Accuracy
 & 28.2\% & 57.8\% & 73.1\% & 26.9\%  \\
 ({\color{red}{2}}-20-{\color{red}{2}}) & Fine-tune privacy
 & 34.7   & 37.1   & 27.5 & 24.0  \\
 & Inference privacy
 & 32.9   & 27.1   & 26.3 & 21.6  \\
  & Communication cost(GB)
 & 2.7 & 2.5 & 42.8 & 2.6\\
\midrule
offline GT
 & Accuracy
  & 29.0\% & 59.2\% & 73.2\% & 27.4\%  \\
  ({\color{red}{2}}-20({\color{red}{8}})-{\color{red}{2}})& Fine-tune privacy
  & N/A & N/A & N/A & N/A  \\
  & Inference privacy
  & 30.4 & 30.1 & 24.6 & 21.3  \\
  & Communication cost(GB)
 & 2.6 & 2.6 & 2.6 & 2.6\\
\bottomrule
\end{tabular}%
\begin{tablenotes}
\item[1] "Fine-tune privacy" and "Inference privacy" values represent the DRA attack effectiveness(ROUGE-L F1) achieved during the respective phases; the lower, the better privacy.
\item[2] "N/A" means the selected privacy attacks are infeasible.
\item[3] The nums below each design name show num of layers for (input adapter-backbone(emulator))-output adapter). Those in red are layers given to the client.
\end{tablenotes}
\end{threeparttable}
\end{table}
\begin{table}[!htbp]
\centering
\scriptsize
\begin{threeparttable}
\caption{Utility and privacy trade-offs on OPT-6.7B.}
\label{tab:opt-6.7b}
\begin{tabular}{lcccccccc}
\toprule
\multirow{2}{*}{Designs} & 
\multirow{2}{*}{Metrics} & 
\multicolumn{4}{c}{OPT-6.7B} \\ 
\cmidrule(lr){3-6}
 & & 
 OpenbookQA & ARC-Easy & PIQA & WebQs \\ 
\midrule
Zero Shot
 & Accuracy
 & 27.6\% & 65.6\% & 76.2\% & 8.8\% \\
\midrule
SL tuning
 & Accuracy
 & 34.9\% & 68.4\% & 78.1\% & 33.9\%  \\
 ({\color{red}{3}}-26-{\color{red}{3}}) & Fine-tune privacy
 & 85.1 & 93.2 & 89.4 & 89.1 \\
 & Inference privacy
 & 83.9 & 92.7 & 85.2 & 87.3 \\
 & Communication cost(GB)
 & 33.6 & 29.8 & 655.5 & 31.3\\
\midrule
offsite tuning
 & Accuracy
 & 33.5\% & 67.2\% & 77.8\% & 30.9\%  \\
 ({\color{red}{3}}-26({\color{red}{14}})-{\color{red}{3}})& Fine-tune privacy
 & N/A & N/A & N/A & N/A \\
 & Inference privacy
 & 100 & 100 & 100 & 100 \\
 & Communication cost(GB)
 & 21.8 & 21.8 & 21.8 & 21.8 \\
\midrule 
online GT
 & Accuracy
 & 33.6\% & 67.8\% & 77.9\% & 31.7\%  \\
 ({\color{red}{3}}-26-{\color{red}{3}})& Fine-tune privacy
 & 46.2 & 46.9 & 27.6 & 39.5 \\
 & Inference privacy
 & 35.6 & 41.7 & 26.3 & 24.2  \\
 & Communication cost(GB)
 & 12.4 & 11.4 & 172.5 & 11.8\\
\midrule
Gradfree GT
 & Accuracy
 & 32.1\% & 66.4\% & 76.9\% & 30.4\%  \\
 ({\color{red}{3}}-26-{\color{red}{3}})& Fine-tune privacy
 & 42.6 & 44.1 & 27.1 & 33.9  \\
 & Inference privacy
 & 41.5 & 42.5 & 27.4 & 32.5  \\
 & Communication cost(GB)
 & 8.7 & 8.2 & 88.7 & 8.4\\
\midrule
offline GT
 & Accuracy
 & 33.2\% & 67.1\% & 77.2\% & 30.1\%  \\
 ({\color{red}{3}}-26({\color{red}{14}})-{\color{red}{3}})& Fine-tune privacy
 & N/A & N/A & N/A & N/A  \\
 & Inference privacy
 & 34.6 & 37.5 & 19.7 & 29.3 \\
 & Communication cost(GB)
 & 16.8 & 16.8 & 16.8 & 16.8\\
\bottomrule
\end{tabular}%
\begin{tablenotes}
\item Please refer to the descriptions in Table~\ref{tab:opt-1.3b} for the same format explanation
\end{tablenotes}
\end{threeparttable}
\end{table}

Both \textit{online} and \textit{Gradfree GuardedTuning} designs send 4 pre-train model layers to the client in OPT-1.3B model experiment, which is the server's model privacy cost.
\textit{Gradfree GuardedTuning} mitigates DRAs by eliminating backpropagation to the server. However, by only training output adapter layers, it incurs a cost of reduced model utility.
\textit{Offline GuardedTuning} provides the strongest client-data privacy by keeping client data entirely local during fine-tuning, also protecting inference privacy through keep the fine-tuned adapters locally during the inference phase. Using the emulator for fine-tuning degrades the model utility slightly compared to \textit{online GuardedTuning} but still better than \textit{Gradfree GuardedTuning}, consistent with observations in offsite-tuning~\cite{xiao2023offsite}.
There are potentially bigger risks to sever model privacy since the client has the emulator generated from the base model.
We leave risk quantification of an emulator for future investigation.

Trade-offs between utility and privacy results on OPT-6.7B are shown in Table \ref{tab:opt-6.7b}.
For these experiments, we opt for a choice that fine-tunes only the adapter layers in both SL tuning and \textit{online GuardedTuning}. This choice leads to computation cost savings, and even with that reduced resource for fine-tuning, all tuned models still achieve utility higher than zero-shot.
We observe similar privacy and utility characteristics to our OPT-1.3B experiments.

%% file: conclusion.tex
\section{CONCLUSION}
In this paper, we address the challenge of producing privacy-preserving fine-tuning designs that balance protecting the client's data privacy, model provider's intellectual property, and the overall tuning cost.  
We introduce several novel designs, collectively named \Workname. 
Through these designs, we showcase how to make design tradeoffs to meet various model utility, privacy, and cost requirements for real-world deployments.
Through experiments, we demonstrate that each of our designs achieves distinct tradeoffs through selective combinations of architecture and novel privacy-enhancement methods, and they all provide defenses against some state-of-the-art privacy attacks on client data.

%% file: main.bbl
\begin{thebibliography}{10}

\bibitem{berant2013semantic}
J.~Berant, A.~Chou, R.~Frostig, et~al.
\newblock Semantic parsing on freebase from question-answer pairs.
\newblock In {\em Proceedings of EMNLP}, pages 1533--1544, 2013.

\bibitem{bisk2020piqa}
Y.~Bisk, R.~Zellers, J.~Gao, et~al.
\newblock Piqa: Reasoning about physical commonsense in natural language.
\newblock In {\em Proceedings of AAAI}, volume~34, pages 7432--7439, 2020.

\bibitem{chen2024unveiling}
G.~Chen, Z.~Qin, M.~Yang, et~al.
\newblock Unveiling the vulnerability of private fine-tuning in split-based frameworks for large language models: A bidirectionally enhanced attack.
\newblock {\em arXiv preprint}, 2024.

\bibitem{clark2018think}
P.~Clark, I.~Cowhey, O.~Etzioni, et~al.
\newblock Think you have solved question answering? try arc, the ai2 reasoning challenge.
\newblock {\em arXiv preprint}, 2018.

\bibitem{nvidia_gpu_secure}
N.~Corporation.
\newblock Confidential computing on nvidia h100 gpus for secure and trustworthy ai.
\newblock \url{https://nvidia.com/h100-secure-ai}, 2023.

\bibitem{deng2021tag}
J.~Deng, Y.~Wang, J.~Li, et~al.
\newblock Tag: Gradient attack on transformer-based language models.
\newblock {\em arXiv preprint}, 2021.

\bibitem{lm-evaluation-harness}
EleutherAI.
\newblock Lm evaluation harness.
\newblock \url{https://bit.ly/LM-EvalHarness}, 2024.

\bibitem{azure_confidential}
Microsoft.
\newblock Confidential containers on azure container instances.
\newblock \url{https://aka.ms/confidential-containers}, 2024.

\bibitem{mihaylov2018can}
T.~Mihaylov, P.~Clark, T.~Khot, et~al.
\newblock Can a suit of armor conduct electricity? a new dataset for open book question answering.
\newblock {\em arXiv preprint}, 2018.

\bibitem{song2020information}
C.~Song and A.~Raghunathan.
\newblock Information leakage in embedding models.
\newblock In {\em Proceedings of CCS}, pages 377--390, 2020.

\bibitem{touvron2023llama}
H.~Touvron, T.~Lavril, G.~Izacard, et~al.
\newblock Llama: Open and efficient foundation language models.
\newblock {\em arXiv preprint arXiv:2302.13971}, 2023.

\bibitem{vepakomma2018split}
P.~Vepakomma, O.~Gupta, T.~Swedish, et~al.
\newblock Split learning for health: Distributed deep learning without sharing raw patient data.
\newblock {\em arXiv preprint}, 2018.

\bibitem{vepakomma2020nopeek}
P.~Vepakomma, A.~Singh, O.~Gupta, et~al.
\newblock Nopeek: Information leakage reduction to share activations in distributed deep learning.
\newblock In {\em Proceedings of ICDMW}, pages 933--942. IEEE, 2020.

\bibitem{wang2023privatelora}
Y.~Wang, Y.~Lin, X.~Zeng, et~al.
\newblock Privatelora for efficient privacy preserving llm.
\newblock {\em arXiv preprint}, 2023.

\bibitem{wei2021finetuned}
J.~Wei, M.~Bosma, V.~Y. Zhao, et~al.
\newblock Finetuned language models are zero-shot learners.
\newblock {\em arXiv preprint}, 2021.

\bibitem{wu2024fedbiot}
F.~Wu, Z.~Li, Y.~Li, et~al.
\newblock Fedbiot: Llm local fine-tuning in federated learning without full model.
\newblock In {\em Proceedings of KDD}, pages 3345--3355, 2024.

\bibitem{xiao2023offsite}
G.~Xiao, J.~Lin, and S.~Han.
\newblock Offsite-tuning: Transfer learning without full model.
\newblock {\em arXiv preprint}, 2023.

\bibitem{zhang2022opt}
S.~Zhang, S.~Roller, N.~Goyal, et~al.
\newblock Opt: Open pre-trained transformer language models.
\newblock {\em arXiv preprint}, 2022.

\end{thebibliography}
